%% file: main.tex
\documentclass[10pt,twocolumn,letterpaper]{article}

\usepackage{iccv}              

\input{preamble}

\definecolor{iccvblue}{rgb}{0.21,0.49,0.74}
\usepackage[pagebackref,breaklinks,colorlinks,allcolors=iccvblue]{hyperref}

\def\paperID{4880} 
\def\confName{ICCV}
\def\confYear{2025}

\let\oldtwocolumn\twocolumn
\renewcommand\twocolumn[1][]{
    \ifshowteaser
        \global\showteaserfalse
        \oldtwocolumn[{#1}{
            \centering
            \vspace{-20pt}
            \includegraphics[width=\textwidth]{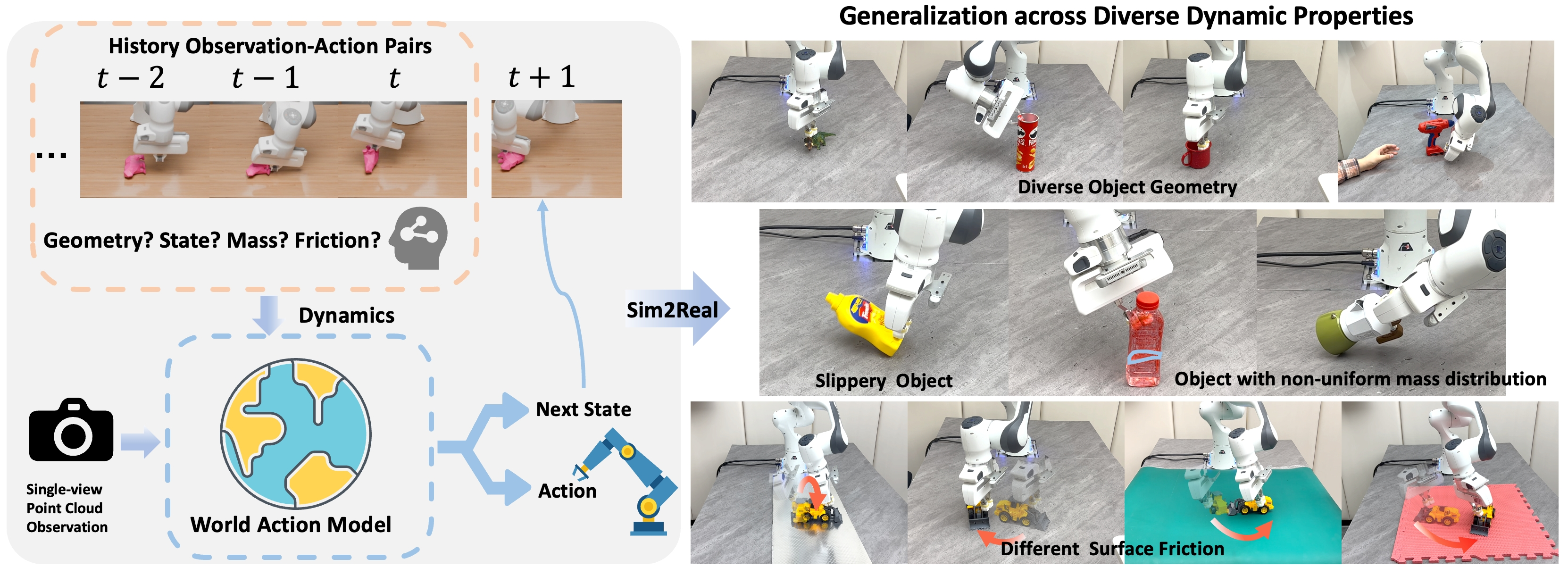}  
            \vspace{-15pt}
            \captionof{figure}{\textbf{Illustration of the high-level idea and generalization ability of DyWA.} Given a target object's 6D pose and \textit{single-view} object point cloud, our non-prehensile manipulation policy aims to rearrange the object without grasping. \textbf{Left}: Our key insight is to enhance action learning by jointly predicting future states while adapting to dynamics from historical trajectories. (For clarity, rendered images are used for visualization, while the actual visual input consists of partial point clouds.)
            \textbf{Right}: After being trained in simulation, our policy achieves zero-shot sim-to-real transfer and generalizes across diverse dynamic properties, including variations in object geometry, object physical property (e.g., slipperiness and non-uniform mass distribution), and surface friction.
            }      
            \vspace{5pt}
            \label{fig:teaser}
        }]
    \else
        \oldtwocolumn[#1]
    \fi
}

\title{DyWA: Dynamics-adaptive World Action Model \\  for Generalizable Non-prehensile Manipulation}

\author{
Jiangran Lyu\textsuperscript{\rm1,\rm2}, 
Ziming Li\textsuperscript{\rm1,\rm2},
Xuesong Shi\textsuperscript{\rm2},
Chaoyi Xu\textsuperscript{\rm2},
Yizhou Wang\textsuperscript{\rm1,\rm3,\rm4,\dag},
He Wang\textsuperscript{\rm1,\rm2,\dag}\\
{\small
\textsuperscript{\rm 1}Center on Frontiers of Computing Studies, School of Computer Science, Peking University \quad
\textsuperscript{\rm 2}Galbot }\\
{\small
\textsuperscript{\rm 3}Inst. for Artificial Intelligence, Peking University \quad
\textsuperscript{\rm 4}State Key Laboratory of General Artificial Intelligence, Peking University
}\\
{\tt\small \url{https://pku-epic.github.io/DyWA/}
}
}

\begin{document}
\maketitle
\input{sec/0_abstract}    
\input{sec/1_intro}
\input{sec/2_related}

\input{sec/3_method}

\input{sec/4_experiments}
\input{sec/5_conclusion}

\section*{Acknowledgements}
We thank Yixin Zheng for organizing the code release, and Junhao Yang for assistance with rendering. We also appreciate the valuable suggestions and discussions from Jiayi Chen, Jiazhao Zhang, Mi Yan and Shenyuan Gao. This work was supported in part by National Science and Technology Major Project (2022ZD0114904) \& NSFC-6247070125.

{
    \small
    \bibliographystyle{ieeenat_fullname}
    \bibliography{main}
}

\input{sec/X_suppl}

\end{document}

%% file: preamble.tex
\usepackage{multirow}
\usepackage{makecell}
\usepackage{pifont}
\usepackage{bbding}
\usepackage{siunitx}
\usepackage{svg}
\usepackage{bbm}

\newif\ifshowteaser
\showteasertrue

%% file: sec/0_abstract.tex
\begin{abstract}
\renewcommand{\thefootnote}{}
\footnote{$\dagger$: Corresponding authors}
Nonprehensile manipulation is crucial for handling objects that are too thin, large, or otherwise ungraspable in unstructured environments. While conventional planning-based approaches struggle with complex contact modeling, learning-based methods have recently emerged as a promising alternative. However, existing learning-based approaches face two major limitations: they heavily rely on multi-view cameras and precise pose tracking, and they fail to generalize across varying physical conditions, such as changes in object mass and table friction.  
To address these challenges, we propose the Dynamics-Adaptive World Action Model (DyWA), a novel framework that enhances action learning by jointly predicting future states while adapting to dynamics variations based on historical trajectories. By unifying the modeling of geometry, state, physics, and robot actions, DyWA enables more robust policy learning under partial observability.
Compared to baselines, our method improves the success rate by 31.5\% using only single-view point cloud observations in the simulation. Furthermore, DyWA achieves an average success rate of 68\% in real-world experiments, demonstrating its ability to generalize across diverse object geometries, adapt to varying table friction, and robustness in challenging scenarios such as half-filled water bottles and slippery surfaces.
\end{abstract}
\vspace{-5mm}

%% file: sec/1_intro.tex
\section{Introduction}
\label{sec:intro}

Non-prehensile manipulation—such as pushing, sliding, toppling, and flipping—greatly extends the capabilities of robotic manipulators beyond traditional pick-and-place operations. These dexterous actions enable robots to handle tasks where grasping is infeasible or inefficient due to object geometry, clutter, or workspace constraints. Over the years, significant progress has been made in this area, particularly through planning-based approaches \cite{mordatch2012contact, posa2014direct,moura2022non,yang2024dynamic}. While effective, these methods typically rely on prior knowledge of object properties, such as mass, friction coefficients, or even complete CAD models, which limits their practicality in real-world applications.
Recently, learning-based methods \cite{zhang2023learning} have emerged as a promising alternative, improving generalization across diverse unseen objects. In this paradigm, policies are trained in simulation and then deployed zero-shot in the real world. For instance, HACMan \cite{zhou2023hacman} leverages vision-based reinforcement learning (RL) on object surface point clouds to determine contact locations and motion directions for executing action primitives. Similarly, CORN \cite{cho2024corn} employs a teacher-student distillation framework, where a teacher policy is first trained using RL with privileged state knowledge and then distilled into a vision-based student policy.

However, these methods face two key limitations that hinder robust real-world deployment.
First, as noted by \cite{ferrandis2024learning}, they rely heavily on multi-view cameras for accurate object geometry and on precise pose tracking modules for state estimation. In practical settings, nevertheless, multi-view setups may be unavailable, and tracking modules are often imperfect, leading to unreliable state information.
Second, these approaches struggle to generalize across diverse physical conditions, such as variations in object mass and table friction, as their models primarily focus on geometry while overlooking the underlying dynamics.

In contrast, we argue that a  generalizable non-prehensile manipulation policy in a realistic robotic setting should not only accommodate diverse object geometries but also adapt to varying physical properties, all while relying solely on a single-camera setup without the need for additional tracking modules.

To achieve this objective, we first experiment with the popular teacher-student policy distillation framework under this challenging setting. Our experiments reveal that while the RL teacher policy, when given oracle information, achieves high performance across diverse dynamic conditions, the distilled student policy, relying on partial observations, suffers from a significant performance drop. We then identify three key factors contributing to this issue.
First, severe partial observability from single-view setting harms action learning by omitting critical geometric cues.
Second, the Markovian student model inherently learns only an averaged behavior across diverse physical variations, resulting in suboptimal performance. 
Third, conventional distillation methods supervise only latent features and final actions, which is insufficient to capture the underlying dynamics necessary for effectively learning contact-rich action. 

To address the first two issues, we introduce a Dynamics Adaptation Module, inspired by RMA \cite{kumar2021rma}, which encodes historical observation-action pairs to model dynamic properties, incorporating both sufficient geometric and physical knowledge.
For the third issue, we extend conventional action learning by enforcing the joint prediction of actions and their corresponding future states. This reformulation transforms the conventional action model into a world action model, introducing additional supervisory signals beyond those provided by the teacher. This synergistic learning paradigm improves imitation loss optimization and significantly enhances overall success rates.
Finally, to guide the world action model with the dynamics embedding adequately, we bridge the two parts using Feature-wise Linear Modulation (FiLM) conditioning. 
In short, we propose a novel policy learning framework that jointly predicting future states while adapting dynamics from historical trajectories. We term our approach \textbf{DyWA (Dynamics-Adaptive World Action Model)}.

We conduct extensive experiments in both simulation and the real world to evaluate the effectiveness and generalization of our policy, comparing it against baseline methods.
To address the lack of a unified benchmark for non-prehensile manipulation, we build a comprehensive benchmark based on CORN, varying camera views (one or three) and the presence of a ground-truth pose tracker.
Our method demonstrates the superiority of its model design across different settings, with a 31.5\% improvement in success rate than baselines. Furthermore, comprehensive ablation studies validate the synergistic benefits of dynamics adaptation and world modeling when jointly learning actions.  
Finally, real-world experiments show that DyWA generalizes across object geometries at a 68\% success rate and adapts to physical variations like table friction. It also achieves robustness in handling non-uniform mass distributions (e.g., half-filled water bottles) and slippery objects. Additionally, we showcase its applications combined with VLM, which assists human or grasping models with thin or wide objects.  

In summary, this work makes the following contributions:  
\begin{itemize}  

    \item We propose DyWA, a novel policy learning approach by jointly predicting future states, with adaptation of dynamics modeling from historical trajectories.
      
    \item We improve generalizable non-prehensile manipulation, reducing dependence on multi-camera setups and pose tracking modules while ensuring robustness across varying physical conditions.
    
    \item We provide a comprehensive simulation benchmark for generalizable non-prehensile manipulation. Our approach surpasses all baseline methods, and we showcase its effectiveness through several real-world applications.
\end{itemize}  

%% file: sec/2_related.tex
\section{Related Works}
\label{sec:related}
\input{figures/pipeline/pipeline}

\subsection{Non-prehensile Manipulation}
Non-prehensile manipulation refers to the process of manipulating objects without grasping them \cite{mason1999progress}. This form of manipulation involves complex contact interactions among the robot, the object, and the environment, posing significant challenges for state estimation, planning, and control \cite{cheng2022contact,hou2019robust,mordatch2012discovery,yu2016more}.
A line of prior work has employed planning-based approaches, either by relaxing contact mode decision variables \cite{mordatch2012contact} or by introducing complementarity constraints to manage contact mode transitions \cite{posa2014direct,moura2022non,yang2024dynamic}.
However, these planners typically assume prior knowledge of object properties, such as mass, friction coefficients, or even complete CAD models, which limits their practical applicability. 

In contrast, learning-based methods \cite{zhang2023learning} offer a promising alternative by enabling generalization without relying on known physical parameters. Nonetheless, most existing works are constrained either by the complexity of the manipulation skills—such as being limited to 2D planar pushing \cite{wu2020spatial,zeng2018learning,pinto2017learning,ferrandis2024learning, ding2024preafford}—or by limited generalization across diverse object types \cite{liang2023learning,zhou2023learning,kim2023pre}.
Recent advances employing point cloud observations, such as HACMan and its variants \cite{zhou2023hacman,le2024enhancing,jiang2024hacman++}, as well as CORN \cite{cho2024corn}, have demonstrated the feasibility of 6D object rearrangement, capturing more complex object interactions while generalizing to a wide range of unseen geometries. However, as highlighted by \cite{ferrandis2024learning}, these methods rely on multi-view cameras and object state estimation through accurate pose tracking modules.
In contrast, our method achieves accurate end-to-end 6D non-prehensile manipulation with generalization across diverse object geometry, varying physics condition using only one single-view camera in the real world. 

\subsection{World Models}
World models \cite{ha2018world} learn compact representations of the environment and predict future states conditioned on action sequences. They have been widely studied in domains such as gaming \cite{ha2018recurrent, hafner2019dream, hafner2020mastering, sekar2020planning} and autonomous driving \cite{wang2024driving, hu2022model, chen2024drivinggpt}. 

In robotic manipulation, jointly modeling future observations and actions has shown strong empirical performance \cite{cheang2024gr, yang2023learning, zhu2025uwm, wuunleashing, guo2024prediction}. We adopt the term \emph{world action model} to characterize this class of policy learning methods with world model attribute, distinguishing it from prior work such as JOWA \cite{cheng2024scaling}.

%% file: figures/pipeline/pipeline.tex
\begin{figure*}[t]
    \centering
    \includegraphics[width=0.95\linewidth]{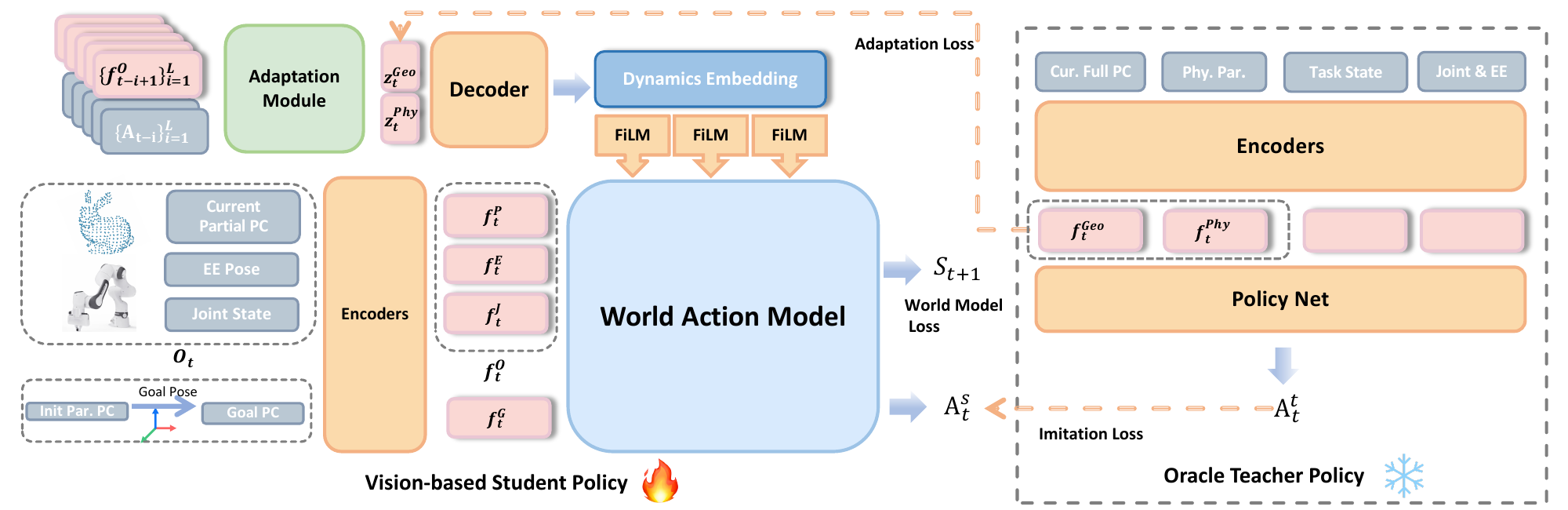}
    \caption{Our World Action Model processes the embeddings of the current observation (partial point cloud, end-effector pose, and joint state) and the goal point cloud (transformed from the initial partial observation) to predict the robot action and next state. Additionally, an adaptation module encodes historical observations and actions, decoding them into the dynamics embedding that conditions the model via FiLM. A pre-trained RL teacher policy (right) supervises both the action and adaptation embedding using privileged full point cloud and physics parameter embeddings. }
    \vspace{-2mm}
	\label{fig:pipeline}
\end{figure*}

%% file: sec/3_method.tex
\section{Method}

\subsection{Task Formulation}
Following HACMan and CORN, we focus on the task of 6D object rearrangement via non-prehensile manipulation. The robot's objective is to execute a sequence of non-prehensile actions (\ie, pushing, flipping) to move an object on the table to a target 6D pose.  
We define the goal pose $\mathbf{G}$ as a 6DoF transformation relative to the object's initial pose, assuming both are stable on the table. The task state $S_t$ at timestep $t$ is represented by the relative transformation between the object's current pose and the goal pose.  
Observations include the partial point cloud $P_t$, joint states $J_t$, and end-effector pose $E_t$. 

\subsection{Pipeline Overview}  
Our training pipeline follows a standard teacher-student distillation framework. Due to the difficulty of obtaining high-quality demonstrations for our task, we first train a state-based RL policy with additional privileged information—\ie, the full object point cloud, task state, and physical parameters—as the teacher policy.  
For consistency, we adopt the same reward design as CORN, as elaborated in the supplementary material. To obtain a vision-based policy suitable for real-world deployment, we introduce our Dynamics-adaptive World Action Model, which serves as the student policy distilled from the teacher policy. Unlike the teacher, our student model relies solely on limited observations that are feasible to obtain in real-world settings.  

In the following sections, we detail the design of the world action model (Sec.~\ref{sec:world_action_model}) and the dynamics adaptation mechanism (Sec.~\ref{sec:adaptation}). To enable adaptive force interaction in this contact-rich manipulation, we further incorporate a variable impedance controller (Sec.~\ref{sec:impedance_control}). Once trained (Sec.~\ref{sec:learning_protocol}), our model can be transferred from simulation to the real world in a zero-shot manner, without requiring real-world fine-tuning.

\subsection{World Action Model}
\label{sec:world_action_model}
\paragraph{Definition.} 
A \emph{world action model} refers to a policy model that jointly predicts actions and forecasts the corresponding future states. Although the current action is not provided as an explicit input, the model exhibits world model characteristics by implicitly conditioning on the current policy action prediction.

\paragraph{Observation and Goal Encoding.}  
Our model takes observation and goal description as input, encoding different modalities using individual encoders.  
For the partial point cloud observation, we process it using a simplified PointNet++ \cite{qi2017pointnet++} to obtain $\mathbf{f}_t^P$. The architectural details are provided in the supplementary material.  
For robot proprioception, we separately encode joint positions and velocities ($\mathbf{f}_t^J$) and the end-effector pose ($\mathbf{f}_t^E$) using shallow MLPs.  
For the Goal Description, instead of relying on the unknown task state $S_t$, we construct a visual goal representation by transforming the initial point cloud $P_0$ to the goal pose, yielding $P_G = \mathbf{G} P_0$. This goal point cloud is then encoded using the shared network with the observation point cloud encoder.  

\paragraph{State-based World Modeling.}  
We enforce the end-to-end model that jointly makes action decisions and predicts their outcomes, creating a synergistic learning process that, in turn, improves action learning.  
Specifically, the observation and goal embeddings are processed through MLPs to produce both the action $\mathbf{A}_t$ and the next task state $S_{t+1}$, with supervision signals separately derived from the teacher policy and simulation outcomes.  
Our object-centric world model represents the environment using task state $S_{t+1}$ instead of high-dimensional visual signals, enabling the policy to focus on task-relevant dynamics. To represent rotations, we adopt the 9D representation \cite{levinson2020analysis, lyuscissorbot}, and define the world model loss as:  
\begin{equation}
\mathcal{L}_{\text{world}} = \| \mathbf{T}_{t+1} - \hat{\mathbf{T}}_{t+1} \|_2^2 + \| \mathbf{R}_{t+1} - \hat{\mathbf{R}}_{t+1} \|_1
\end{equation}
where $\mathbf{T}_{t+1} \in \mathbb{R}^3$ and $\mathbf{R}_{t+1} \in SO(3)$ are the predicted translation and rotation, while $\hat{\mathbf{T}}_{t+1} \in \mathbb{R}^3$ and $\hat{\mathbf{R}}_{t+1} \in SO(3)$ denote the ground-truth transformation obtained from simulation outcomes after action execution.  
Additionally, we employ an imitation loss, defined as the L2 loss between the predicted action and the teacher action:  
\begin{equation}
    \mathcal{L}_\text{imitation} = \|\mathbf{A}_t^s - \mathbf{A}_t^t \|^2  
\end{equation}  

\subsection{Dynamics Adaptation} 
\label{sec:adaptation}
To enhance the world model's ability to adapt to diverse dynamics, we extract abstract representations of environmental variations from historical trajectories.  
Our approach distills teacher knowledge regarding full point cloud and physical parameter into an adaptation embedding, which is subsequently decoded into the dynamics embedding. This embedding then conditions the world action model through a learnable feature-wise linear modulation mechanism.  

\paragraph{Adaptation Embedding.}  
We design an adaptation module that processes sequential observation-action pairs to compensate for missing geometry
and physics knowledge in the current partial observation. Specifically, at each timestep, we concatenate the observation embeddings $f_t^O = \{f_t^P, f_t^J, f_t^E\}$ with the previous action embedding $f_{t-1}^A$, where the action embedding is obtained via a shallow MLP.  
We construct an input sequence of $L$ past observation-action tuples which is then processed by a 1D CNN-based adaptation module, for extracting a compact adaptation embedding:  
\begin{equation}
\mathbf{z}_t = \mathrm{Embed}\left( \left[ \mathrm{concat}(\mathbf{f}^O_{t-i-1}, \mathbf{f}^A_{t-i-2}) \right]_{i=1}^{L} \right)
\end{equation}
To ensure meaningful representation learning, we supervise the adaptation embedding using the concatenation of the full point cloud embedding and physics embedding from the teacher encoder.  
\begin{equation}
    \mathcal{L}_\text{adapt} =  \| \mathbf{z}_t^{Geo, Phy} - concat(\mathbf{f}_t^{Geo}, \mathbf{f}_t^{Phy}) \|^2 
\end{equation}

\paragraph{Dynamics Conditioning.}  
Once the adaptation embedding is obtained, we decode it into the dynamics embedding, which serves as a conditioning input for the world action model via Feature-wise Linear Modulation (FiLM).  
FiLM \cite{perez2018film} dynamically modulates the intermediate feature representations of the world action model by applying learned scaling and shifting transformations, allowing the model to adapt to varying dynamics.  
Each FiLM block consists of two shallow MLPs which take the dynamics embedding as input and output the modulation parameters $\gamma$ and $\beta$ for each latent feature $f$:  
\begin{equation}
    \text{FiLM}(f | \gamma, \beta) = \gamma f + \beta
\end{equation}  
We integrate FiLM blocks densely in the early layers of the world action model while leaving the final layers unconditioned.  
The technique that has proven highly effective in integrating language guidance into vision encoders \cite{chi2023diffusion,brohan2022rt}. In our case, this mechanism allows the dynamics embedding to selectively influence feature representations, enabling adaptive adjustments to the model’s behavior based on the underlying dynamics.  

\subsection{Action Space with Variable Impedance} 
\label{sec:impedance_control}
To enable adaptive force interaction between the robot and object, we employ variable impedance control \cite{cho2024corn} as the low-level action execution mechanism. This allows the robot to dynamically regulate the interaction force based on task demands.  
Specifically, the action space of our policy consists of the subgoal residual of the end effector, $\Delta T_{ee} \in SE(3)$, along with joint-space impedance parameters.  
The joint-space impedance is parameterized by positional gains ($P \in \mathbb{R}^7$) and damping factors ($\rho \in \mathbb{R}^7$), where the velocity gains are computed as $D = \rho \sqrt{P}$.  
To execute the commanded end-effector motion, we first solve for the desired joint position using inverse kinematics with the damped least squares method \cite{buss2004introduction}:  
\begin{equation}
    q_d = q_t + IK(\Delta T_{ee})
\end{equation}  
Then, the desired joint position $q_d$ and impedance parameters $K, D$ are applied to a joint-space impedance controller to generate impedance-aware control commands for the robot. We utilize the widely adopted Polymetis API \cite{Polymetis2021} for implementation.  

\subsection{Training Protocol}  
\label{sec:learning_protocol}
The overall learning objective is formulated as the sum of the imitation loss, world model loss, and adaptation loss:  
\begin{equation}
    \mathcal{L} = \mathcal{L}_\text{imitation} + \mathcal{L}_\text{world} + \mathcal{L}_\text{adapt}
\end{equation}  
We begin by training the teacher policy for 200K iterations in simulation using PPO. Subsequently, we employ DAgger \cite{ross2011reduction} to train the student policy under teacher supervision for 500K iterations.  
To enhance robustness and generalization, we introduce domain randomization during training by varying the object's mass, scale, and friction, as well as the restitution properties of the object, table, and robot gripper. The object scale is adjusted such that its largest diameter remains within a predefined range.  
To further improve sim-to-real transfer, we inject small perturbations into the torque commands, object point cloud, and goal pose when training the student policy.

%% file: sec/4_experiments.tex
\section{Experiments}

\input{tables/exp_main}

\input{tables/exp_ablation}

\input{tables/exp_real}

\input{tables/exp_friction}

\input{figures/loss/loss}

\subsection{Benchmarking Tabletop Non-prehensile Rearrangement in Simulation}  
We evaluate our method alongside several baselines within a unified simulation environment to enable a fair comparison of their performance.  
Although prior works \cite{cho2024corn, zhou2023hacman} have developed their own simulation environments for training and validating non-prehensile manipulation policies, there remains a lack of a standardized benchmark for evaluating both existing and future approaches. To bridge this gap, we establish a comprehensive benchmark based on the CORN setting.  
Specifically, we adopt the IsaacGym simulation environment and utilize 323-object asset from DexGraspNet \cite{wang2023dexgraspnet} for training. Additionally, we enrich the task setting by introducing an unseen object test set, consisting of 10 geometrically diverse objects, each scaled to five different sizes, resulting in a total of 50 evaluation objects.  
Furthermore, we introduce two additional perception dimensions: (i) single-view vs. multi-view (three-camera) observations and (ii) whether known object poses for constructing the task state $S_t$. Both the training and testing environments are fully randomized \wrt dynamics properties including mass, friction, and restitution.

\paragraph{Task Setup.}  
At the beginning of each episode, we randomly place the object in a stable pose on the table. The robot arm is then initialized at a joint configuration uniformly sampled within predefined joint bounds, positioned slightly above the workspace to prevent unintended collisions with the table or object.  
Next, we sample a random 6D stable goal pose on the table, ensuring it is at least $\SI{0.1}{\metre}$ away from the initial pose to prevent immediate success upon initialization. To guarantee valid initial and goal poses for each object, we precompute a set of stable poses, as detailed in the supplementary.  
An episode is considered successful if the object's final pose is within $\SI{0.05}{\metre}$ and \SI{0.1} radians of the target pose.

\paragraph{Baselines.}  
We evaluate our approach against two state-of-the-art baselines: HACMan and CORN, which represent primitive-based and closed-loop methods, respectively. Since HACMan was originally implemented in the MuJoCo simulator, we re-implemented it within our benchmark for a fair comparison. However, because it requires strict per-point correspondence as input, its success rate is extremely low in the unknown state setting. 
CORN shares the same simulation environment as our method, allowing us to train and evaluate it directly with minimal modifications.  
To ensure a fair comparison, we further enhanced CORN by replacing its shallow MLP-based point cloud encoder with the same vision backbone as ours. Additionally, for settings where the current object pose is unknown, we provided all methods with the same goal point cloud representation to maintain consistency.  

\paragraph{Results.}  
As shown in Table \ref{tab:sim}, our method consistently outperforms all baselines across all three evaluation tracks. In particular, we achieve a significant performance gain over previous approaches, with at least a \textbf{31.5\%} improvement in success rate. Notably, the performance gap is most pronounced in challenging scenarios involving unknown states and single-view observations, where our method’s dynamics modeling capability plays a crucial role.  
Compared to HACMan, our approach benefits from its closed-loop execution and variable impedance control, enabling more robust dexterous manipulation. While HACMan relies on pre-defined motion primitives, its adaptability to complex geometries and variations in physics are limited. 
Moveover, our method surpasses CORN due to our adaptation mechanism refines the world model based on historical trajectories, allowing the policy to adjust effectively to variations in object properties such as mass, friction, and scale.  
These results highlight the effectiveness of our strong generalization capabilities in diverse rearrangement tasks.  

\subsection{Ablation Study}  
We conduct ablation studies on the most challenging evaluation track, \ie, unknown state with single-view observation. Our goal is to systematically analyze the contribution of each key module to the overall performance.  

\paragraph{Synergy between Next State Prediction and Action Learning.}
To analyze the optimization process, we visualize the loss curve during training and compare the approach that uses only dynamics adaptation (i.e., RMA) with that adding World Modeling. Our results show that during the distillation, simultaneous learning of the next state improves action loss convergence, confirming the synergy between world modeling and action learning. Additionally, we discuss the integration of the world model in the RL teacher policy, which is elaborated in the supplementary material.

\paragraph{On the Complementarity of Dynamics Adaptation and World Modeling.}  
We investigate the individual and combined effects of dynamics adaptation and world modeling. Our results (Table \ref{tab:ablation}) show that using only the world model or dynamics adaptation, \ie RMA, provides only marginal improvements over the naive DAgger baseline, with success rates increasing by just 1.7\% and 5.7\%, respectively. However, when both modules are used together, the performance jumps significantly from 59.9\% to 73.3\%.  
This improvement can be attributed to the complementary nature of these components. Without dynamics adaptation, the world model lacks sufficient information to reason about the dynamic effects of interaction. Conversely, using only dynamics adaptation also provides limited benefits due to the absence of a sufficiently structured learning target. These findings highlight the complementarity of world modeling and dynamics adaptation, demonstrating that their combination is a non-trivial yet highly effective design choice.  

\paragraph{Effectiveness of FiLM Conditioning.}  
We further evaluate the role of Feature-wise Linear Modulation (FiLM) in bridging adaptation embeddings and the world action model. Our results indicate that FiLM provides a more effective and structured conditioning mechanism than direct input concatenation. Specifically, incorporating FiLM into RMA boosts performance from 65.6\% to 70.0\%. More notably, when all three modules (world modeling, dynamics adaptation, and FiLM) are used together, the success rate reaches 82.2\%, with FiLM contributing an additional 8.9\% improvement.  
We also discuss different methods for conditioning in the supplementary whose conclusion consists with our claims.
This reinforces FiLM as a lightweight and effective choice for integrating adaptation embeddings.

\input{figures/real_demo/real_demo}
\input{figures/vlm/vlm}
\input{figures/pregrasping/pregrasping}

\subsection{Real-World Experiments}  
To evaluate the real-world applicability of our method, we conduct experiments on a physical robot setup. Our goal is to validate the zero-shot transferability of our policy from simulation to the real world and compare its performance against prior methods.  

\paragraph{Real-World Setup}  
Our experimental setup is illustrated in the supplementary. We use a Franka robot arm for action execution and a RealSense D435 camera positioned at a side view to capture RGB-D images. We evaluate our approach on 10 unseen real-world objects, including both slippery objects and those with non-uniform mass distribution such as a half-filled bottle. 
Before each episode, we first place the object at the target goal pose and record its point cloud. Then, we reposition the object in a random stable pose and allow our policy to execute the manipulation task. Upon completion, we use Iterative Closest Point (ICP) to measure the pose error between the final object position and the recorded target pose. For symmetric objects where direct ICP alignment is ambiguous, we relax the success criteria along the symmetric axes and compute errors only in translation and relevant rotational components.


\paragraph{Generalization across Diverse Objects.}
We evaluate our model's generalization ability by comparing it with CORN, which relies on an external tracking module for object pose estimation in real-world experiments. As shown in Figure \ref{fig:real_demo} and Table \ref{tab:real}, our method achieves accurate manipulation across diverse objects without external pose tracking, significantly outperforming CORN with an average success rate of 68\% versus 36\%. CORN struggles with precise execution due to occlusions in single-view partial point clouds and inaccuracies in real-world pose estimation. Additionally, our model demonstrates robust performance on slippery objects and those with non-uniform mass, where CORN fails.
We validate the generalization ability of our model and compare our method against CORN, which depends on an external tracking module to estimate object poses in real-world experiments. 

\paragraph{Robustness to Surface Friction Variations.}  
To assess the effectiveness of dynamics adaptation, we conduct experiments on surfaces with varying friction coefficients. We select four tablecloths (Figure \ref{fig:teaser}) with progressive friction levels, \ie $\mu_1, \mu_2,\mu_3,\mu_4$  and use the bulldozer toy as the test object. Additionally, we report the average execution time for successful episodes.  
As shown in Table \ref{tab:friction}, the model without dynamics adaptation exhibits significant performance degradation when interacting with surfaces of different friction levels, leading to erratic execution times. In contrast, our policy with dynamics adaptation maintains consistent success rates while ensuring stable execution times across all surface conditions. This highlights the robustness of our approach in handling diverse real-world contact dynamics.  

\subsection{Applications}  

We present a practical manipulation system that integrates Vision-Language Models (VLMs), our non-prehensile policy, and a grasping model \cite{shi2024asgrasp}. By leveraging VLMs, our goal-conditioned policy can be executed based on natural language instructions. Specifically, we utilize SoFar \cite{qi2025sofar}, a model capable of generating semantic object poses from language commands, to specify goals for our policy. As shown in Figure \ref{fig:vlm}, given the command \textit{“Put the grip of the electric drill into a person's hand”}, SoFar generates the target transformation of the drill (e.g., rotation $\Delta \theta =122^\circ$ and translation $\Delta x, \Delta y = [0.54, 0.09]$), which is then used as the goal for our policy. This enables natural, instruction-driven object handovers, highlighting the potential of our approach in human-robot interaction.

Additionally, we showcase the system outperforms or complements traditional prehensile manipulation. As illustrated in the third row of Figure \ref{fig:real_demo}, a standard pick-and-place strategy struggles to flip a tiny switch due to gripper-table collisions, whereas our policy enables efficient rearrangement in a single continuous motion. Furthermore, our policy serves as an effective pre-grasping step in the system. As shown in Figure \ref{fig:pregrasping}, certain objects are inherently difficult to grasp due to their geometry—for example, a thin card lying flat on a surface or a broad cracker box exceeding the gripper’s maximum span. Our system can firstly reorient these objects into grasp-friendly configurations, significantly improving grasp success rate.

%% file: tables/exp_main.tex
\begin{table*}[t]
    \centering
    {
    \begin{tabular}{llcccccc}
    \toprule
        \multicolumn{1}{l}{\multirow{2}{*}{\textbf{Methods}}} & \multicolumn{1}{l}{\multirow{2}{*}{\textbf{Action Type}}} & \multicolumn{2}{c}{\textbf{Known State (3 view)}} & \multicolumn{2}{c}{\textbf{Unknown State (3 view)}} & \multicolumn{2}{c}{\textbf{Unknown State (1 view)}}\\ 
        \cmidrule(lr){3-4} \cmidrule(lr){5-6} \cmidrule(lr){7-8}
        ~ & ~ & Seen & Unseen & Seen & Unseen & Seen & Unseen \\ \midrule
        HACMan \cite{zhou2023hacman} &Primitive & 3.8(42.2) & 5.7(39.4) & 3.0(23.6) &  4.1(26.5) & 1.5(17.9) & 2.9(18.3) \\ \midrule
        CORN \cite{cho2024corn} &Closed-loop & 86.8 & 79.9 &46.0 &47.8 & 29.0 &29.8   \\ 
        CORN (PN++) &Closed-loop  &87.3  &84.3 &76.1 &75.7  &50.7  &49.4  \\ 
        Ours &Closed-loop & \textbf{87.9} & \textbf{85.0} &\textbf{85.8} &\textbf{82.3} & \textbf{82.2} & \textbf{75.0} \\ 
        \bottomrule
    \end{tabular}}
    \vspace{-3mm}
    \caption{Quantitative results measured by success rate in the simulation benchmark. For HACMan, we also reports its performance given 3 DoF planar goal(\ie $[\Delta x, \Delta y, \Delta \theta]$) in parentheses. Note that the third track with unknown state and single view camera is the most realistic and challenging track for fully comparison of each methods.}
    \label{tab:sim}
    \vspace{-3mm}
\end{table*}

%% file: tables/exp_ablation.tex
\begin{table}[t]
    \centering
    {
    \begin{tabular}{lccccc}
    \toprule
        Methods & W.M. & D.A. & FiLM & Seen & Unseen \\ \midrule
        DAgger \cite{ross2011reduction} &\ding{56} &\ding{56} &\ding{56} & 59.9 &57.5   \\ 
        World Model &\Checkmark &\ding{56} &\ding{56} & 61.6 &59.4  \\ 
        RMA \cite{kumar2021rma} &\ding{56} &\Checkmark &\ding{56} & 65.6 & 57.9  \\ 
        Ours w/o W.M. &\ding{56} &\Checkmark &\Checkmark & 70.0  & 63.7  \\ 
        Ours w/o FiLM &\Checkmark &\Checkmark &\ding{56} & 73.3 & 59.4  \\ 
        Ours &\Checkmark &\Checkmark &\Checkmark & \textbf{82.2} & \textbf{75.0} \\ 
        \bottomrule
    \end{tabular}}
    \caption{Ablation study on the most challenging evaluation track, \ie, unknown state with single-view observation. W.M. means World Model and D.A. means Dynamics Adaptation.}
    \label{tab:ablation}
\end{table}

%% file: tables/exp_real.tex
\begin{table*}[t]
    \centering
    \resizebox{2\columnwidth}{!}
    {
    \begin{tabular}{lccccccccccc}
    \toprule
        \multicolumn{1}{c}{\multirow{2}{*}{Methods}}  & \multicolumn{7}{c}{Normal} & \multicolumn{1}{c}{Slippery} & \multicolumn{2}{c}{Non-uniform Mass} &\multicolumn{1}{c}{\multirow{2}{*}{Avg.}}\\ 
        \cmidrule(lr){2-8} \cmidrule(lr){9-9} \cmidrule(lr){10-11} 
        &Mug & Bulldozer &Card &Book & Dinosaur & Chips Can &Switch &YCB-Bottle &Half-full Bottle &Coffee jar \\ \midrule
        CORN w tracking &1/5  &3/5  &\textbf{4/5}  &\textbf{4/5} &2/5 &0/5 &2/5 &0/5 &0/5 &2/5 &  18/50 (36\%) \\  
        Ours  &\textbf{3/5}  &\textbf{4/5}  &\textbf{4/5}  &\textbf{4/5} &\textbf{3/5} & \textbf{2/5} &\textbf{4/5} &\textbf{3/5} &\textbf{4/5} & \textbf{3/5} & \textbf{34/50 (68\%)}\\ 
        \bottomrule
    \end{tabular}}
    \vspace{-3mm}
    \caption{Quantitative results in the real world. Each cell shows the number of successful trials out of 5 attempts. Our method consistently achieves high success rates across diverse objects.}
    \label{tab:real}
\end{table*}

%% file: tables/exp_friction.tex
\begin{table*}[t]
    \centering
    {
    \begin{tabular}{lccccccccccc}
    \toprule
        \multicolumn{1}{c}{\multirow{2}{*}{Methods}}  & \multicolumn{2}{c}{$\mu_1$} & \multicolumn{2}{c}{$\mu_2$} & \multicolumn{2}{c}{$\mu_3$} &\multicolumn{2}{c}{$\mu_4$}
        \\ 
        \cmidrule(lr){2-3}  \cmidrule(lr){4-5} \cmidrule(lr){6-7} \cmidrule(lr){8-9}
        &S.R. $\uparrow$ & Avg. Time $\downarrow$ &S.R. $\uparrow$ & Avg. Time $\downarrow$ &S.R. $\uparrow$ & Avg. Time $\downarrow$ &S.R. $\uparrow$ & Avg. Time $\downarrow$\\ \midrule
        Ours w/o D.A. &3/5 &\textbf{\textcolor{red!80!black}{65 s}}  &3/5 &\textbf{\textcolor{red!80!black}{81 s}}  &\textbf{4/5} & \textbf{\textcolor{red!80!black}{96 s}} &3/5 &\textbf{\textcolor{red!80!black}{124 s}}  \\  
        Ours  &\textbf{4/5} &\textbf{\textcolor{ForestGreen}{45 s}} &\textbf{4/5} &\textbf{\textcolor{ForestGreen}{50 s}} &\textbf{4/5} &\textbf{\textcolor{ForestGreen}{ 49 s}} &\textbf{4/5} &\textbf{\textcolor{ForestGreen}{51 s}}  \\ 
        \bottomrule
    \end{tabular}}
    \vspace{-3mm}
    \caption{Experiments on different surface friction, with progressive friction levels, $ \mu_1 \textless \mu_2 \textless \mu_3 \textless \mu_4$.}
    \label{tab:friction}
\end{table*}

%% file: figures/loss/loss.tex
\begin{figure}[t]
    \centering
    \begin{tabular}{ll}
    \includegraphics[width=0.46\linewidth]{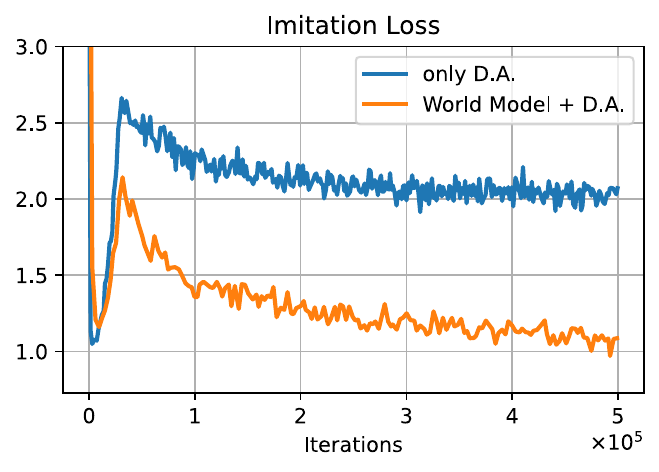} & \hspace{-3mm}
    \includegraphics[width=0.48\linewidth]{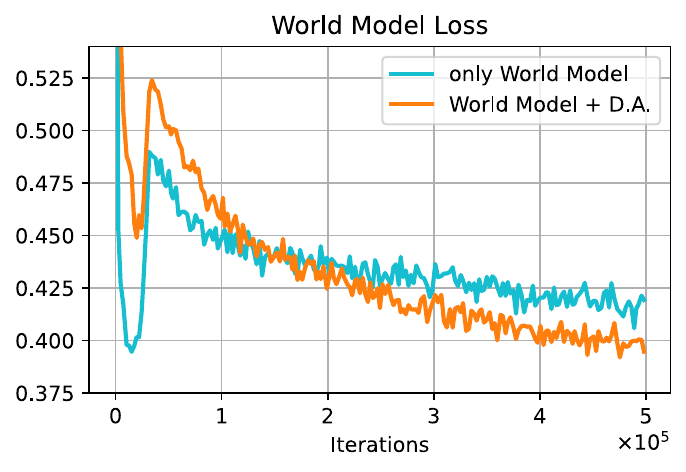}
    \end{tabular}
    \caption{\textbf{Loss curves during the distillation process.} We adopt DAgger which starts with teacher action for execution and gradually adds the weights of student action so that the initial loss declines rapidly. \textbf{Left:} Comparison of imitation loss between using only Dynamics Adaptation and incorporating the World Model. \textbf{Right:} Comparison of World Model loss between using only the World Model and integrating Dynamics Adaptation.}
    \vspace{-2mm}
	\label{fig:loss}
\end{figure}

%% file: figures/real_demo/real_demo.tex
\begin{figure*}[t]
    \centering
    \begin{tabular}{c}
    
    \includegraphics[width=0.85\linewidth]{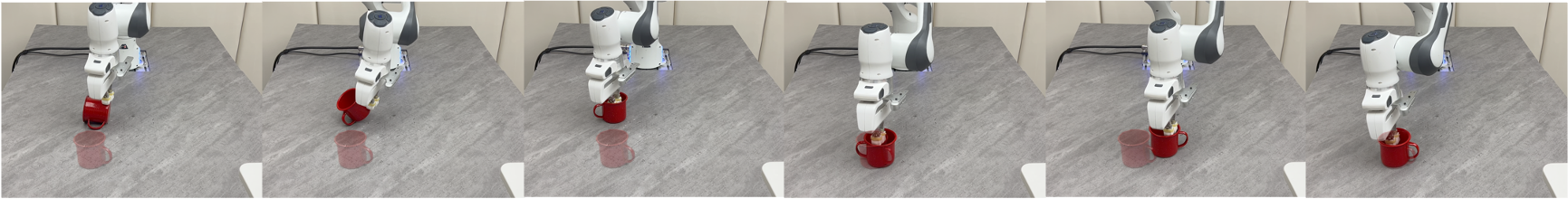} \\
    \includegraphics[width=0.85\linewidth]{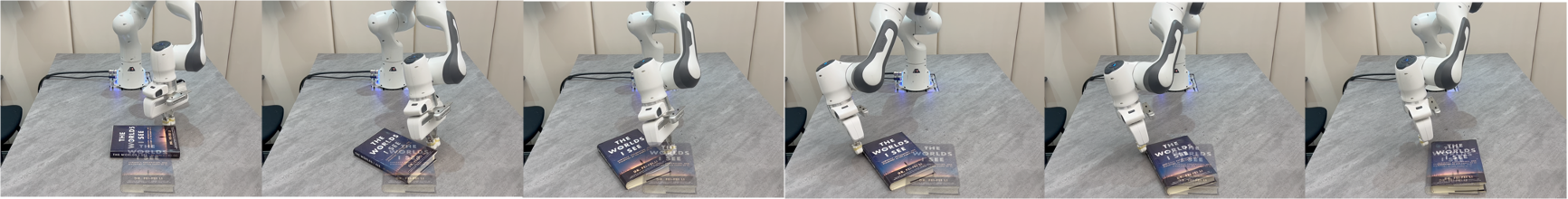} \\
 \includegraphics[width=0.85\linewidth]{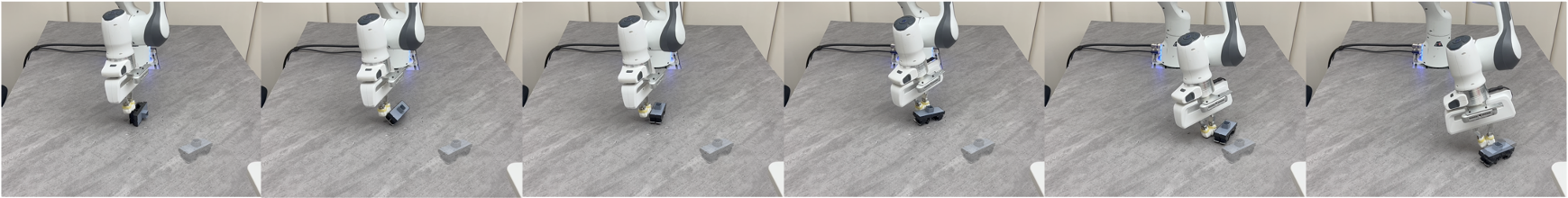}
    
    \end{tabular}
    \vspace{-2mm}
    \caption{Qualatative Results in the real world. The goal pose is shown transparently. }
    \vspace{-2mm}
    \label{fig:real_demo}
\end{figure*}

%% file: figures/vlm/vlm.tex
\begin{figure}[t]
    \centering
    \includegraphics[width=\linewidth]{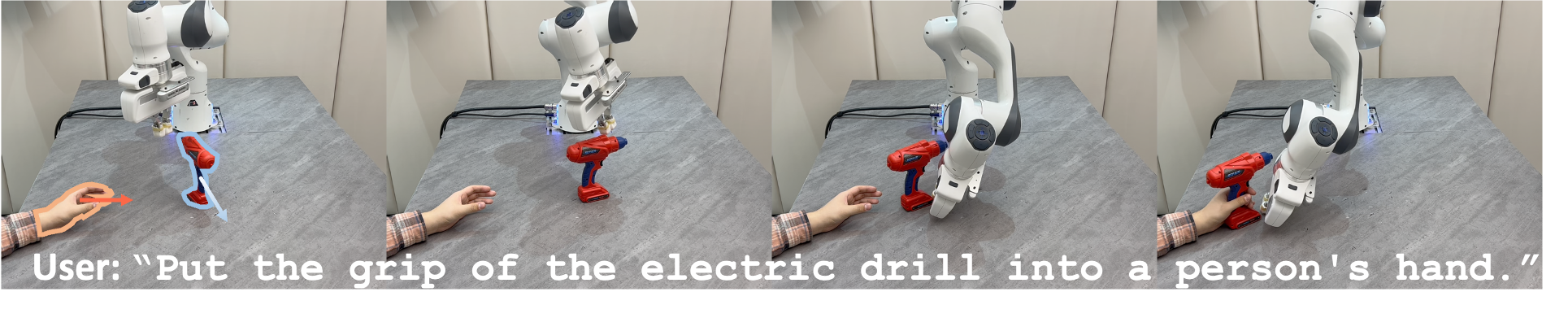}
    \caption{By integrating with Vision-Language Models (VLMs), our goal-conditioned policy can be executed based on natural language instructions. }
    \vspace{-2mm}
	\label{fig:vlm}
\end{figure}

%% file: figures/pregrasping/pregrasping.tex
\begin{figure}[t]
    \centering
    \vspace{-2mm}
    \includegraphics[width=0.95\linewidth]{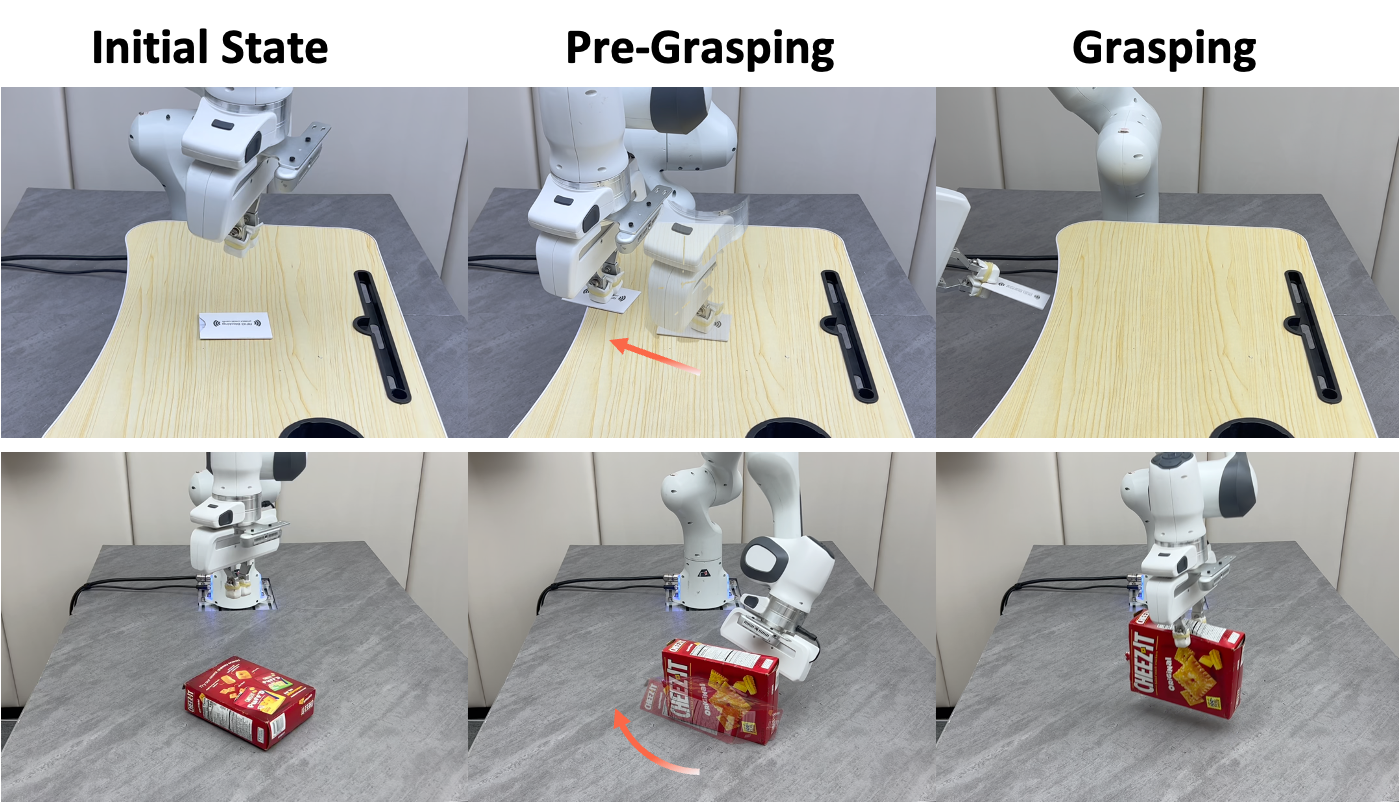}
    \caption{Our policy helps grasping a thin card and broad cracker box.}
    \vspace{-2mm}
	\label{fig:pregrasping}
\end{figure}

%% file: sec/5_conclusion.tex


\section{Conclusion, Limitations, and Future Works}

In this work, we present a novel policy learning approach that jointly predicts future states while adapting dynamics from historical trajectories. Our model enhances generalizable non-prehensile manipulation by reducing reliance on multi-camera setups and pose tracking modules while maintaining robustness across diverse physical conditions.
Extensive simulation and real-world experiments validate the effectiveness of our approach.
However, our method also has certain limitations since it relies solely on point clouds as the visual input modality. It struggles with symmetric objects due to geometric ambiguity, and faces challenges with transparent and specular objects, where raw depth is incomplete. A promising direction is to incorporate additional appearance information \cite{ma2024followyouremoji,ma2025followyourclick,ma2025followcreation,ma2025followyourmotion,ma2023magicstick,ma2024followpose,ma2022visual} to provide richer visual cues. 

%% file: sec/X_suppl.tex
\clearpage
\setcounter{page}{1}
\maketitlesupplementary

\section{Real-world Setup}
\input{figures_supp/real_setup/real_setup}
As shown in Figure \ref{fig:rel_setup}, We use a franka panda robot arm for execution and a Intel RealSense D435 camera for capturing point cloud. We also adopt the same object segmentation method with CORN \cite{cho2024corn}, consists of color-based segmentation followed by depth-based back-projection to obtain a single-view object point cloud. 

\section{RL-based Teacher Policy}
\subsection{Reward Design}  

Following~\citet{kim2023pre, cho2024corn}, the reward function in our domain is defined as:  
\begin{equation}  
r = r_{suc} + r_{reach} + r_{contact} - c_{energy},  
\end{equation}  
where $r_{suc}$ is the task success reward, $r_{reach}$ is the goal-reaching reward, $r_{contact}$ is the contact-inducing reward, and $c_{energy}$ is the energy penalty.  

The task success reward is defined as:  
\begin{equation}  
r_{suc} = \mathbbm{1}_{suc},  
\end{equation}  
where $\mathbbm{1}_{suc}$ is an indicator function that returns 1 when the object’s pose is within 0.05m and 0.1 radians of the target pose.  

To facilitate learning, we introduce dense rewards $r_{reach}$ and $r_{contact}$, formulated based on a potential function~\citep{ng1999policy} as:  
\begin{equation}  
r = \gamma\phi(s') - \phi(s),  
\end{equation}  
where $\gamma \in [0,1)$ is the discount factor. Specifically,  
\begin{equation}  
\phi_{reach}(s) = k_{g} \gamma^{k_{d} \cdot d_{o,g}(s)},  
\end{equation}  
\begin{equation}  
\phi_{contact}(s) = k_{r} \gamma^{k_{d} \cdot d_{h,o}(s)},  
\end{equation}  
where $k_g, k_d, k_r \in \mathbb{R}$ are scaling coefficients. The term $d_{o,g}(s)$ represents the distance between the current object pose and the goal pose, measured using a bounding-box-based distance metric, while $d_{h,o}(s)$ denotes the distance between the object's center of mass and the tip of the gripper.  

The energy penalty is defined as:  
\begin{equation}  
c_{energy} = k_e \sum_{i=1}^{7} \tau_i \dot{q}_i,  
\end{equation}  
where $k_e \in \mathbb{R}$ is a scaling coefficient, and $\tau_i$ and $\dot{q}_i$ denote the torque and velocity of the $i^{\text{th}}$ joint, respectively.

\subsection{Architecture}

\input{tables_supp/exp_teacher}
Our teacher policy architecture consists of separate encoders for each modality and an MLP-based policy network, which proves sufficiently effective for RL training. We also experiment with incorporating our proposed world modeling and FiLM into the teacher policy, as shown in Table \ref{tab:teacher}. However, the performance remains nearly unchanged compared to the baseline approach.
Together with the analysis in Figure \ref{fig:loss}, we attribute this to the RL policy already reaching its performance upper bound given the current architecture. The core contribution of DyWA primarily benefits the distillation process, facilitating better optimization of the imitation loss rather than improving the teacher policy itself.

\section{Alternative of Dynamics Factor Conditioning}
\input{tables_supp/exp_film}
To investigate alternative conditioning mechanisms, we experimented with cross-attention layers as a replacement for FiLM. However, this approach led to significantly degraded performance. We hypothesize that the transformer-based cross-attention mechanism is more sensitive to data distribution shifts and may require additional architectural modifications or extensive data augmentation, introducing unnecessary overhead for this task. These findings further support FiLM as a lightweight yet effective method for integrating adaptation embeddings into the world-action model. 

\section{Simulation Assets and Setup}
\input{figures_supp/sim_setup/sim_setup}
\input{figures_supp/asset/train_set}
Our simulation setup is shown as Figure \ref{fig:sim_setup}.
We sample 323 objects from the DexGraspNet dataset as training set and 10 objects as test set, as shown in Figure \ref{fig:train_set}, \ref{fig:test_set}. 
To sample stable poses for training, we drop the objects in a simulation and extract the poses after stabilization. In 80\% of the trials, we drop the object $\SI{0.2} \meter$ above the table in a uniformly random orientation. In the remaining 20\%, we drop the objects from their canonical orientations, to increase the chance of landing in standing poses, as they are less likely to occur naturally. If the object remains stationary for 10 consecutive timesteps, and its center of mass projects within the support polygon of the object, then we consider the pose to be stable. We repeat this process to collect at least 128 initial candidates for each object, then keep the unique orientations by pruning equivalent orientations that only differ by a planar rotation about the z-axis.

\input{tables_supp/hyper}

\input{tables_supp/hyper_network}

\input{tables_supp/hyper_teacher}

\input{tables_supp/hyper_teacher_network}

\input{figures_supp/sim_demo/sim_demo}

\section{Vision Encoder}
We use a simplified version of PointNet++ \cite{qi2017pointnet++} as our point cloud encoder. 
Specifically, the student policy encoder employs two layers of fixed-scale grouping, while the teacher policy encoder uses one layer. In the $\mathbf{i^{th}}$ grouping layer, $\mathbf{C_i}$ key points are selected via farthest point sampling (FPS), and each key point forms a group with its $\mathbf{K}$ nearest neighbors (KNN). Each cluster is then processed by two per-point MLP layers and two global MLP layers to generate a group feature. The output point cloud consists only of the $\mathbf{C_i}$ selected key points, each enriched with its corresponding $\mathbf{M_i}$-dimensional group feature. After the grouping layers, the per-point features are concatenated and passed through residual MLP layers, following the structure of PointNet++. The final output consists of several point tokens with grouped features.

\section{Hyper-parameters}

The following Tables \ref{tab:hyper_student}, \ref{tab:hyper_studnet_network}, \ref{tab:hyper_teacher_network}, \ref{tab:hyper_teacher} demonstrate the hyper-parameters of our policy and the teacher policy.

%% file: figures_supp/real_setup/real_setup.tex
\begin{figure}[t]
    \centering
    \includegraphics[width=0.95\linewidth]{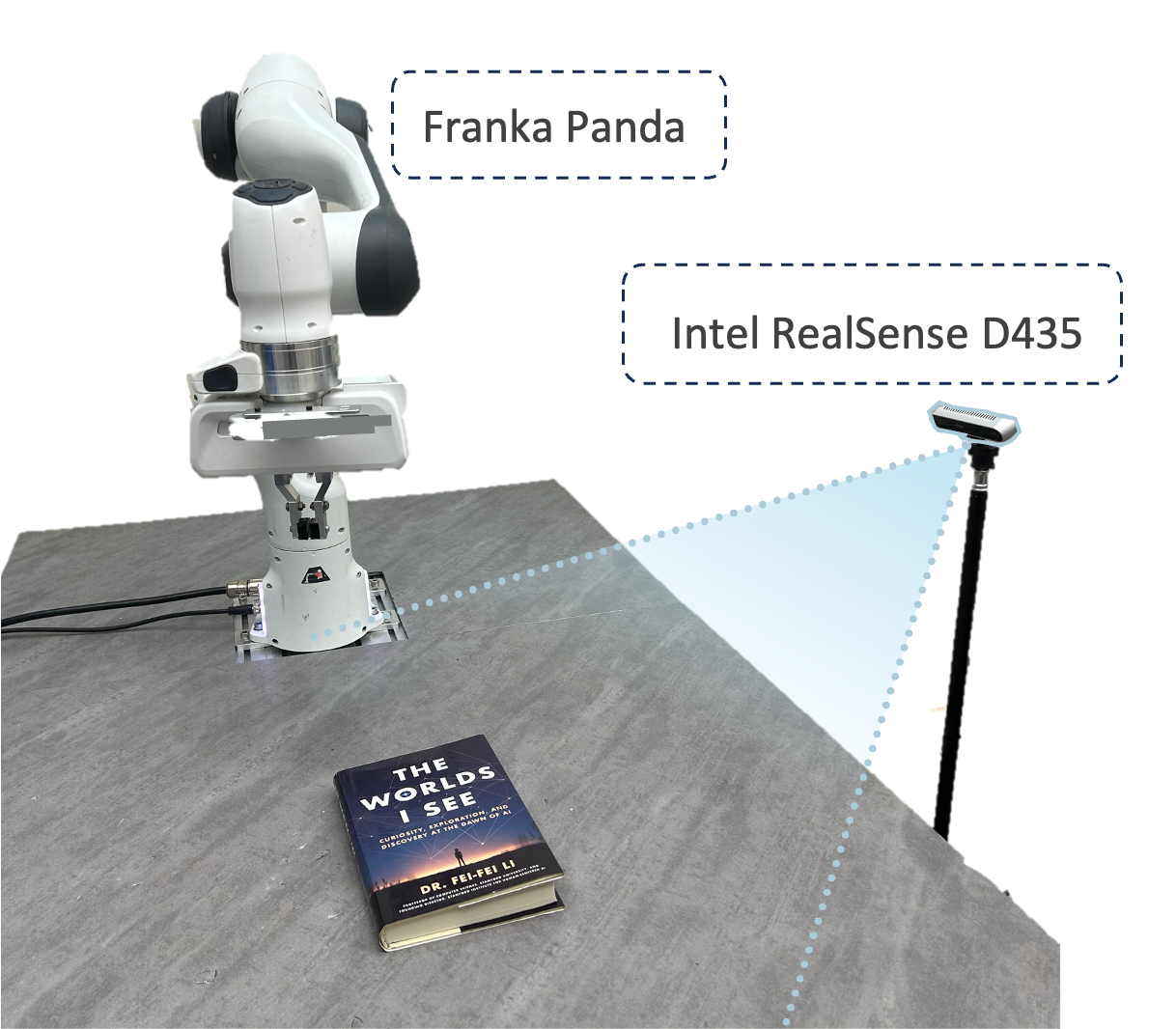}
    \caption{Real-world Experiment Setup}
    \vspace{-2mm}
	\label{fig:rel_setup}
\end{figure}

\begin{figure}[t]
    \centering
    \includegraphics[width=\linewidth]{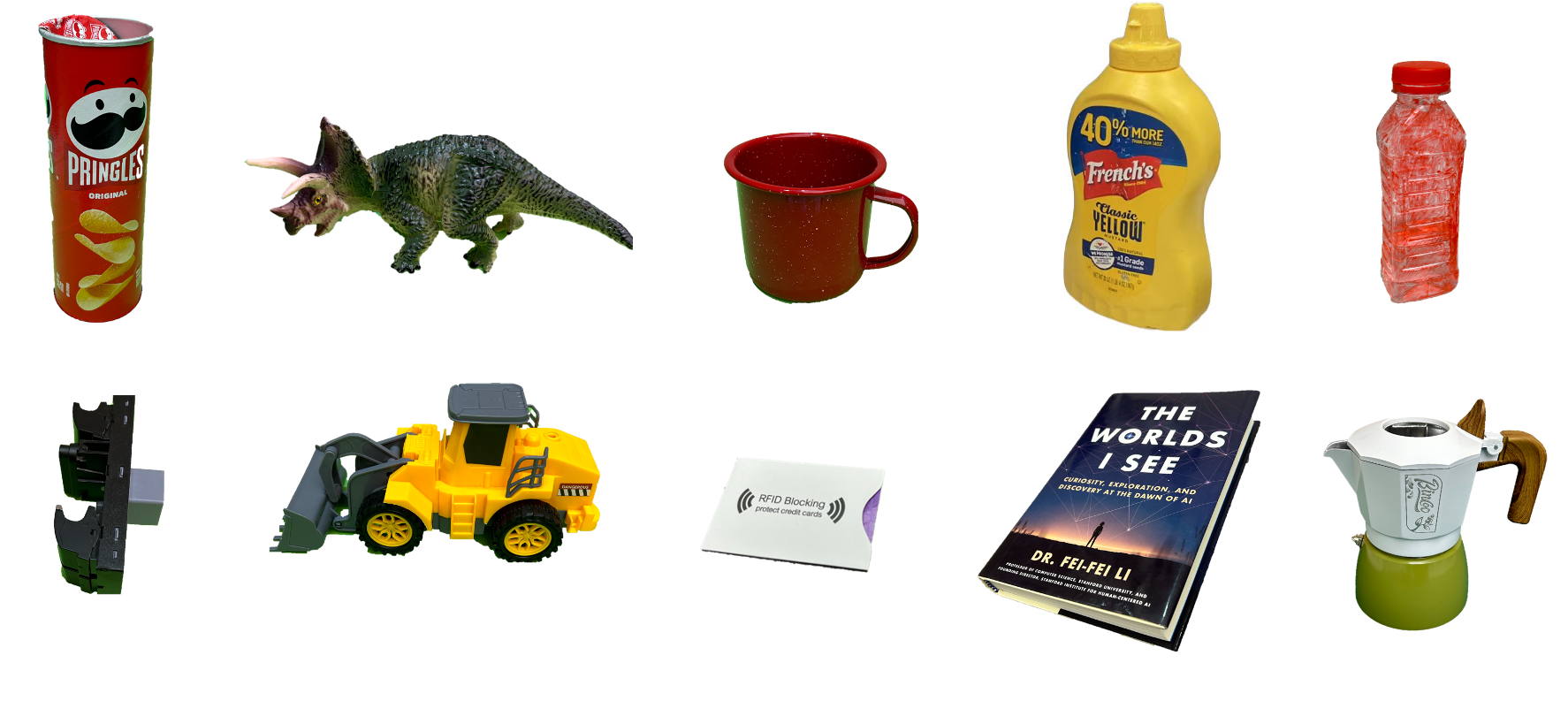}
    \caption{Objects used in the real-world experiments}
    \vspace{-2mm}
	\label{fig:real_objects}
\end{figure}

%% file: tables_supp/exp_teacher.tex
\begin{table}[h]
    \centering
    {
    \begin{tabular}{ccc}
    \toprule
    World Model & FiLM & Success Rate \\ \midrule
    \ding{56} &\ding{56} & 94.1 \\ 
    \Checkmark &\ding{56} & 93.9 \\ 
    \Checkmark &\Checkmark & 93.5 \\ 
    \bottomrule
    \end{tabular}}
    \caption{Success rate of RL-based Teacher Policy.}
    \label{tab:teacher}
\end{table}

%% file: tables_supp/exp_film.tex
\begin{table}[h]
    \centering
    {
    \begin{tabular}{cccc}
    \toprule
        & MLP & Cross Atten & FiLM \\ \midrule
        Success Rate &73.3 &70.1 & 82.2 \\ 
        \bottomrule
    \end{tabular}}
    \caption{Success Rate of student policy with different dynamics conditioning methods.}
    \label{tab:film}
\end{table}

%% file: figures_supp/sim_setup/sim_setup.tex
\begin{figure}[t]
    \centering
    \includegraphics[width=\linewidth]{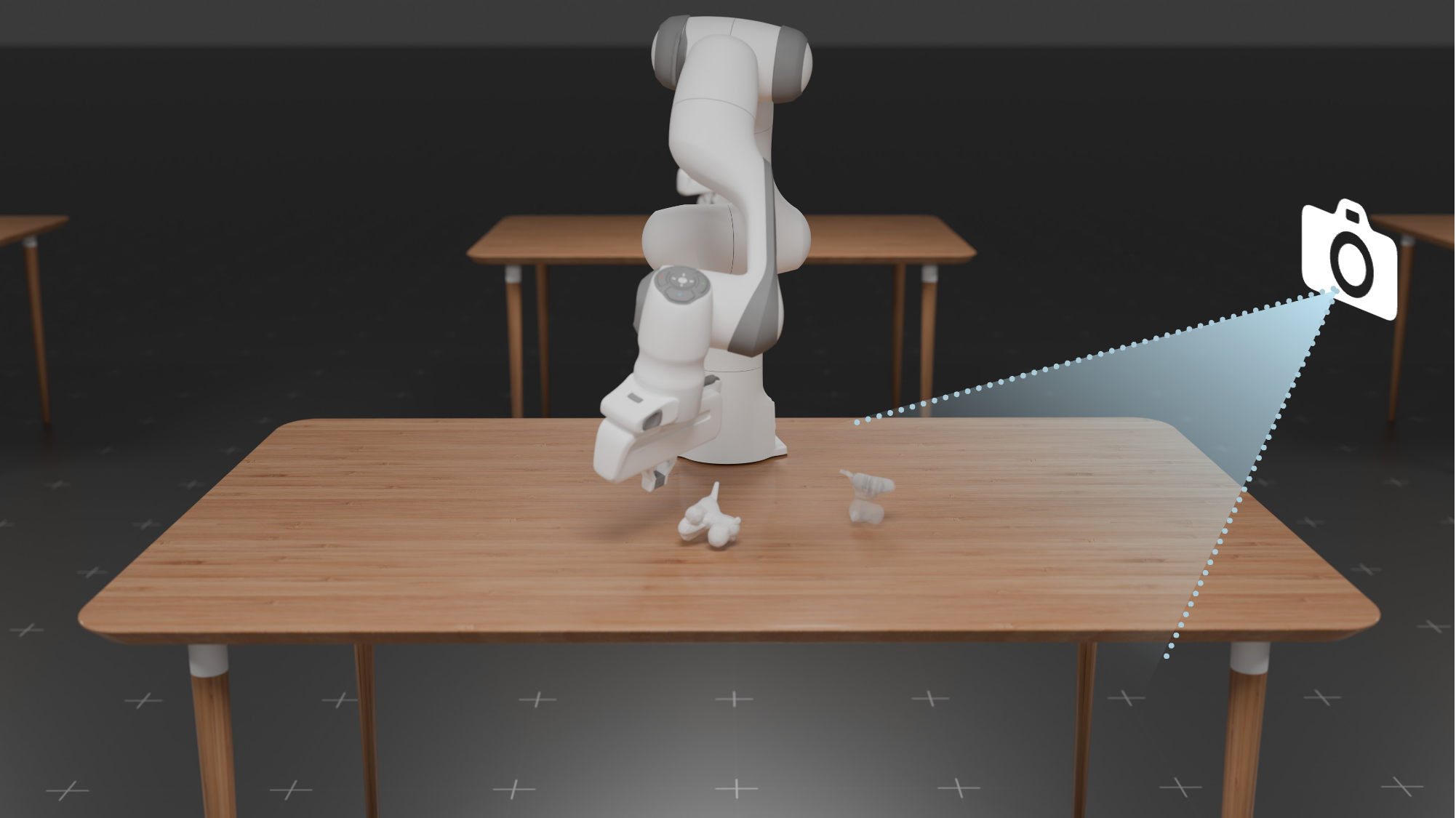}
    \caption{Simulation Environment Setup. }
    \vspace{-2mm}
	\label{fig:sim_setup}
\end{figure}

%% file: figures_supp/asset/train_set.tex
\begin{figure*}[t]
    \centering
    \includegraphics[width=0.95\linewidth]{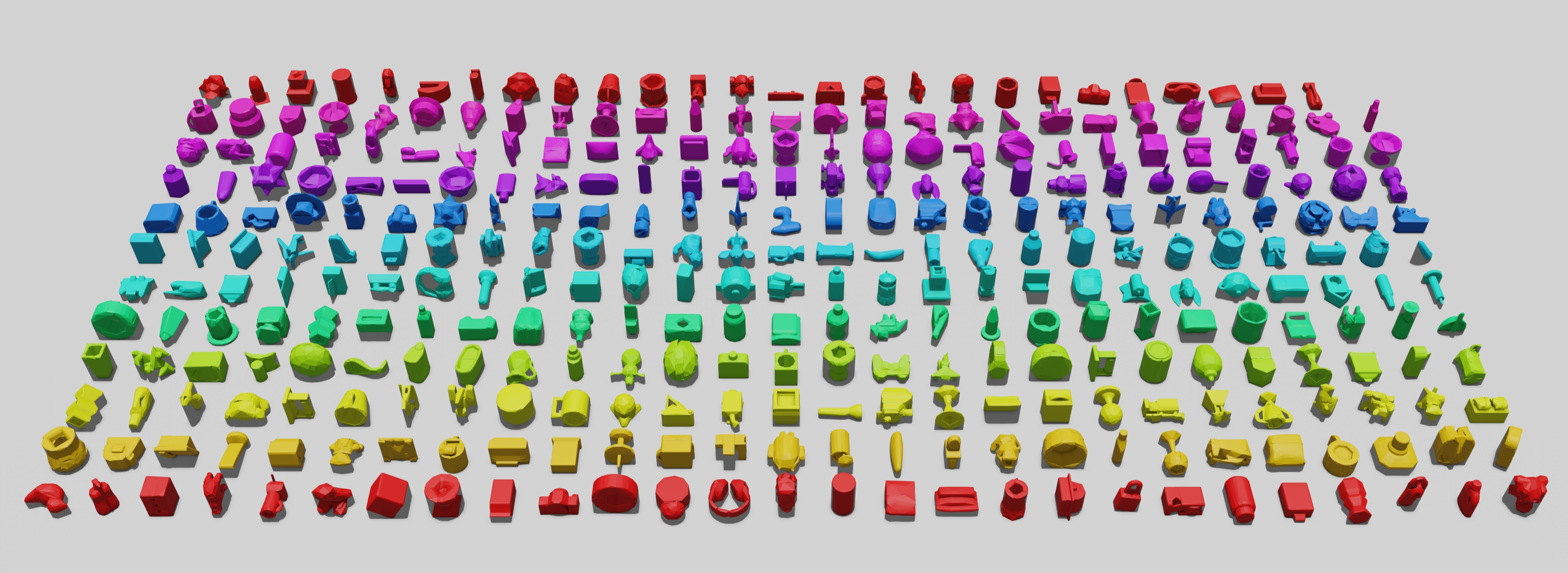}
    \caption{Objects used for training in the simulation benchmark.}
    \vspace{-2mm}
	\label{fig:train_set}
\end{figure*}

\begin{figure}[t]
    \centering
    \includegraphics[width=\linewidth]{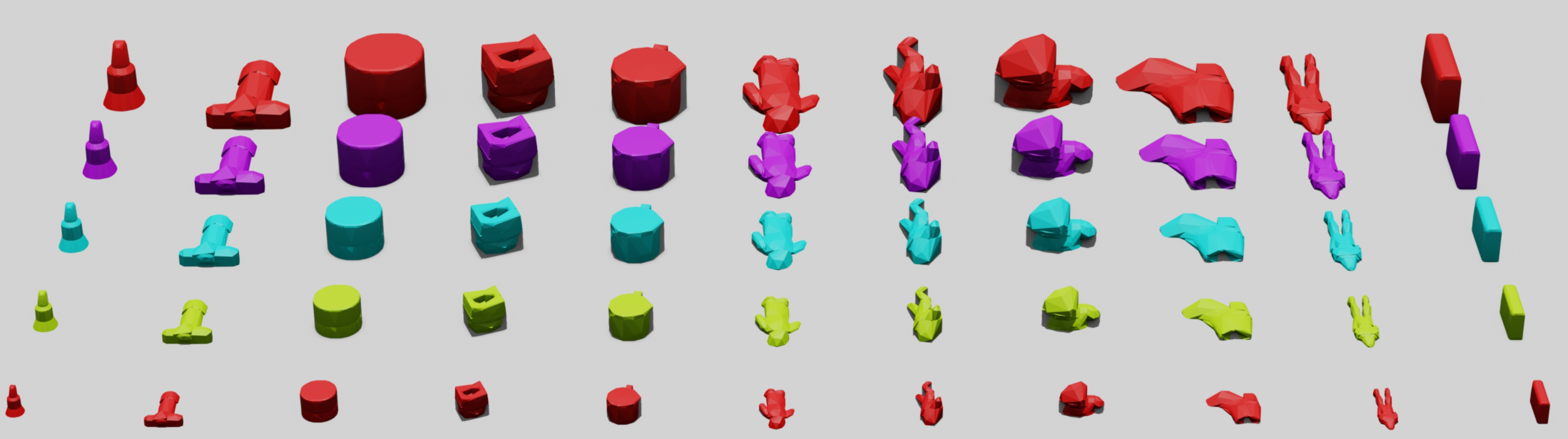}
    \caption{Unseen objects used for evaluation in the simulation benchmark.}
    \vspace{-2mm}
	\label{fig:test_set}
\end{figure}

%% file: tables_supp/hyper.tex
\begin{table}[h]
    \centering
    {
    \begin{tabular}{ll}
    \toprule
    Hyperparameter & Value  \\ \midrule
    Learning rate &  6e-4 \\ 
    Num. Environment & 1024 \\ 
    Optimizer & Adam \\
    Normalization & Layernorm \\
    Dropout & 0 \\
    \bottomrule
    \end{tabular}}
    \caption{Hyper-parameters for Student's Training Algorithm}
    \label{tab:hyper_student}
\end{table}

%% file: tables_supp/hyper_network.tex
\begin{table}[h]
    \centering
    {
    \begin{tabular}{ll}
    \toprule
    Hyperparameter & Value  \\ \midrule
    Input Size & (512, 3) \\
    Key points $\mathbf{C_i}$ &  [64, 16]\\ 
    Grouping Neighbours $\mathbf{K}$ &  32 \\ 
    Grouped feature $\mathbf{M_i}$ & [32, 128] \\ 
    Global points MLP  & MLP(4096, 1024, 1024, 4096) \\
    History length & 5 \\
    History Decoder & Conv1d+MaxPool \\
    History Decoder channel & 128 \\
    FiLM block Num& 3 \\
    Pose Predictor Shared  & MLP(4096, 256) \\
    Translation predictor & MLP(256, 128, 64, 3) \\
    Rotation predictor  & MLP(256, 128, 64, 3) \\
    Actor & MLP(4736, 1024, 256) \\
    \bottomrule
    \end{tabular}}
    \caption{Hyper-parameters for Student's Encoder and Policy}
    \label{tab:hyper_studnet_network}
\end{table}

%% file: tables_supp/hyper_teacher.tex
\begin{table}[h]
    \centering
    {
    \begin{tabular}{ll}
    \toprule
    Hyperparameter & Value  \\ \midrule
    RL algorithm & PPO \\
    Adam stepsize &  3e-4 \\ 
    Num. Environment & 4096 \\ 
    GAE parameter & 0.95 \\
    Discount Factor & 0.99 \\
    PPO clip range & 0.3 \\
    Num. epoch & 8 \\
    \bottomrule
    \end{tabular}}
    \caption{Hyper-parameters for Teacher's PPO Algorithm}
    \label{tab:hyper_teacher_network}
\end{table}

%% file: tables_supp/hyper_teacher_network.tex
\begin{table}[h]
    \centering
    {
    \begin{tabular}{ll}
    \toprule
    Hyperparameter & Value  \\ \midrule
    Key points $\mathbf{C_i}$ & [16] \\
    Grouping Neighbors $\mathbf{K}$ &  32 \\ 
    Grouped feature Channels $\mathbf{M_i}$ & [128] \\ 
    Shared MLP & MLP(512, 256, 128) \\
    Actor & MLP(64, 20) \\
    Critic & MLP(64, 1) \\
    
    \bottomrule
    \end{tabular}}
    \caption{Hyper-parameters for Teacher's Encoder and Policy}
    \label{tab:hyper_teacher}
\end{table}

%% file: figures_supp/sim_demo/sim_demo.tex
\begin{figure*}[t]
    \centering
    \includegraphics[width=0.95\linewidth]{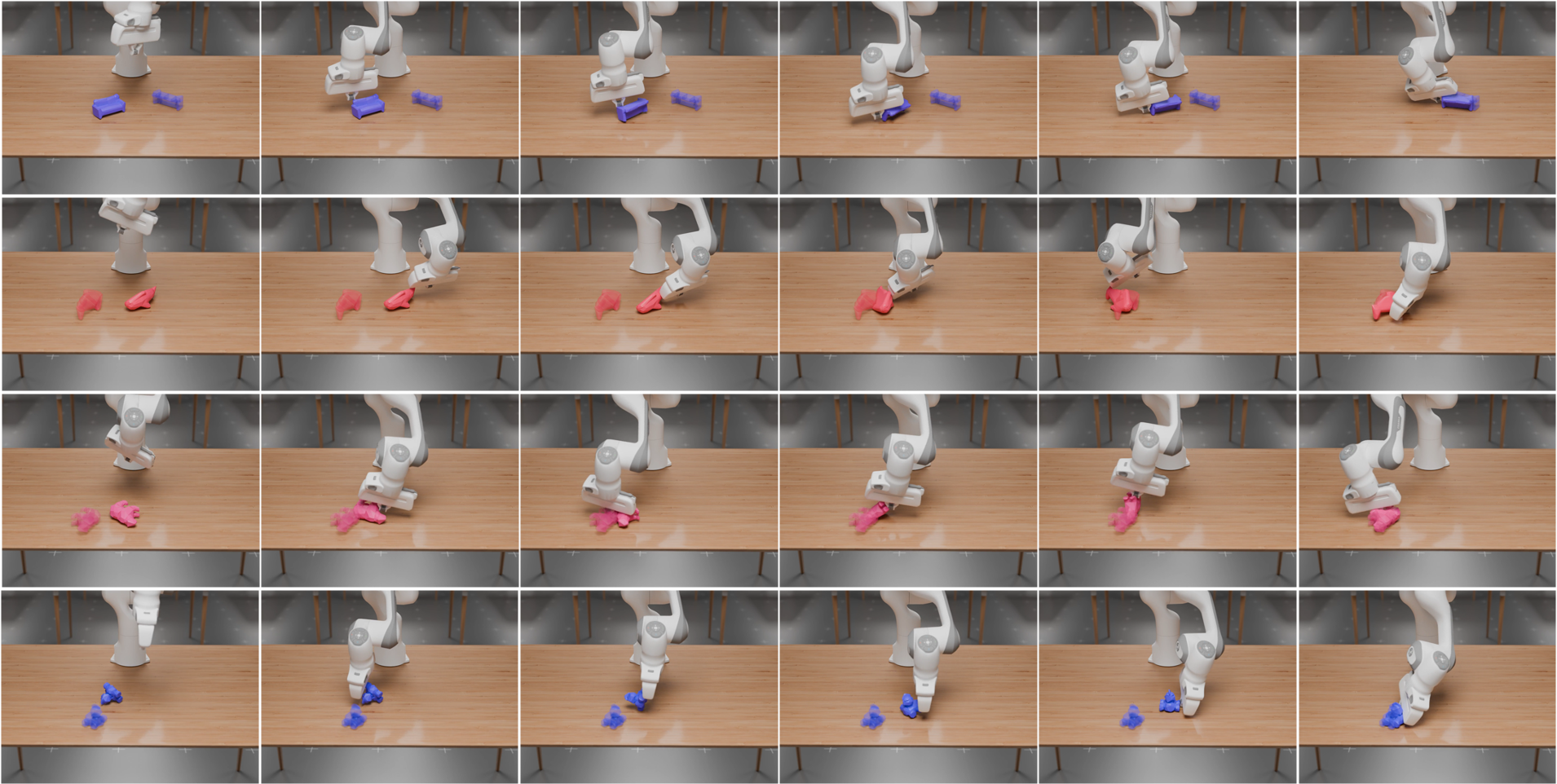}
    \caption{Qualatative results in the simulation.}
    \vspace{-2mm}
	\label{fig:sim_demo}
\end{figure*}